\PassOptionsToPackage{table,dvipsnames}{xcolor}
\documentclass{sjtu_sai}
\usepackage[utf8]{inputenc}
\usepackage[T1]{fontenc}
\usepackage{url}
\usepackage{booktabs}
\usepackage{float}
\usepackage{placeins}
\usepackage{multirow}
\usepackage{pifont}
\usepackage{amsfonts}
\usepackage{nicefrac}
\usepackage{adjustbox}
\usepackage{microtype}
\usepackage[table,dvipsnames]{xcolor}
\usepackage{graphicx}
\usepackage{wrapfig}
\usepackage{amsmath}
\usepackage{amssymb}
\usepackage{longtable}
\usepackage{xspace}
\usepackage{tabularx}
\usepackage{array}
\usepackage{hyperref}
\usepackage[nameinlink,noabbrev]{cleveref}


\newcommand{\PRTI}{\textsc{PRTI}\xspace}
\newcommand{\BenchName}{\textsc{RoboCombine}\xspace}

\newcommand{\NumCompositions}{19\xspace}

\newcommand{\task}[1]{z_{#1}}

\definecolor{lightblue}{RGB}{230,240,255}
\definecolor{lightgray}{RGB}{245,245,245}
\definecolor{claimblue}{RGB}{220,235,255}
\definecolor{sotagreen}{RGB}{222,245,232}
\definecolor{riskred}{RGB}{255,235,232}
\definecolor{refgray}{RGB}{240,240,240}

\newcounter{algobox}

\crefname{algobox}{Algorithm}{Algorithms}
\Crefname{algobox}{Algorithm}{Algorithms}
\newif\ifintrofigureonewrap
\introfigureonewraptrue

\title{Exploring the Limits of Combinatorial Generalization in Embodied Manipulation}

\renewcommand{\authorfont}{\fontsize{11}{14}\selectfont}
\renewcommand{\authorformat}[2][]{\mbox{\authorfont{\bfseries\rmfamily #2$^{#1}$}}}
\author[1]{Xupeng Zhang}
\author[1]{Yantai Yang}
\author[1]{Chang Guo}
\author[1]{Zhaokai Yin}
\author[1,*]{Zhipeng Zhang}
\affiliation[1]{AutoLab, School of Artificial Intelligence, Shanghai Jiao Tong University}
\contribution[*]{Corresponding author.}
\abstract{In embodied manipulation, Vision Language Action (VLA) models can execute isolated atomic skills, but practical long-horizon manipulation requires recombining these skills into new compositions without demonstrations for every skill combination.
The direct \emph{full instruction route} treats skill chaining as long-horizon imitation.
In our controlled study, full instruction training on limited skill combination sequences reaches 40.0\% ordered success on covered combinations but only 7.7\% on recombination, suggesting memorization of sequence templates rather than robust skill recombination.
We therefore introduce \emph{Hierarchical Subtask Execution} (\textsc{HSE}), a diagnostic execution framework that exposes skill recombination through a shared modular interface over the planner, trigger, and policy, enabling long-horizon failures to be decomposed into separable modes.
To make this diagnosis measurable, we introduce \textsc{RoboCombine}, a benchmark for skill recombination without combined skill training trajectories, with metrics for ordered success and transition conversion across subtask boundaries.
\textsc{RoboCombine} reveals that atomic competence alone does not ensure ordered execution.
Failures concentrate at transitions where downstream subtasks must launch from handoff states produced by preceding subtasks rather than canonical atomic starts.
We identify this mismatch as the \emph{start state gap}.
To address it, we introduce \emph{Pose Robust Task Initialization} (\textsc{PRTI}), a lightweight atomic task post-training strategy that broadens start state coverage beyond canonical starts.
Under oracle timed \textsc{HSE} on \textsc{RoboCombine}, \textsc{PRTI} reaches 18.16\% ordered success, compared with 4.21\% under random perturbation and 9.47\% under a robot-reset oracle, without combined skill training trajectories.
Together, these results show that combinatorial generalization requires handoff robust atomic skills that convert first stage progress into reliable downstream execution, not merely decomposition into known subtasks.
\textsc{RoboCombine} and HSE code is available at \url{https://github.com/AutoLab-SAI-SJTU/RoboCombine}.
}

\begin{document}
\maketitle

\section{Introduction}
\label{sec:introduction}

In embodied manipulation, Vision-Language-Action (VLA) models have driven remarkable progress in language-conditioned robotic manipulation, successfully executing various isolated atomic skills such as grasping objects, opening drawers, or moving articulated parts~\citep{brohan2023rt2visionlanguageactionmodelstransfer,kim2024openvlaopensourcevisionlanguageactionmodel,octomodelteam2024octoopensourcegeneralistrobot,black2024pi0visionlanguageactionflowmodel}. However, practical real-world applications demand long-horizon tasks that are fundamentally compositions of these primitives. To achieve broader goals, a generalist robot must sequence multiple operations rather than halting after a single isolated motion.

\begin{wrapfigure}{r}{0.42\textwidth}
    \vspace{-0.8em}
    \centering
    \includegraphics[width=\linewidth]{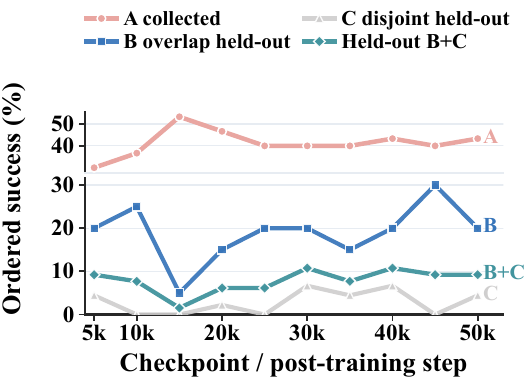}
    \vspace{-15pt}
    \caption{Finite composed trajectory training shows an A$\rightarrow$B$\rightarrow$C degradation as composition novelty increases.}
    \label{fig:intro_combo_ckpt_trend}
    \vspace{-1.1em}
\end{wrapfigure}

One intuitive approach to this challenge is the \emph{full-instruction route}, which treats skill chaining as a long-horizon imitation problem by training models on demonstrations of complete task sequences~\citep{shi2025hirobotopenendedinstruction,yang2025lohovlaunifiedvisionlanguageactionmodel,lin2025onetwovlaunifiedvisionlanguageactionmodel,zhao2025cotvlavisualchainofthoughtreasoning}. While this provides rich supervision, it heavily couples the learned policy to the observed distribution of skill compositions. We address the viability of this route through a controlled investigation. By initializing a VLA trained on atomic skills and continuing full-instruction training on a limited set of composed sequences, we evaluate the model across a spectrum of compositional novelty. As illustrated in Fig.~\ref{fig:intro_combo_ckpt_trend}, we observe a stepwise degradation from seen compositions to the partially overlapping and disjoint sets, with performance averaging only 7.7\% over the two sets. This contrast exposes a critical bottleneck where models easily overfit to finite observed combinations but struggle to transfer to novel ones, suggesting they memorize sequence templates rather than acquiring robust skill recombination capabilities.

To achieve true combinatorial generalization, a fundamentally more scalable direction is the \emph{hierarchical-subtask route}. Instead of executing a composed command as one monolithic instruction, this approach decomposes it into a sequence of atomic subtasks and dynamically updates the VLA conditioning when progress cues indicate a handoff-ready window~\citep{strudel2020learningcombineprimitiveskills,fox2019hierarchicalvariationalimitationlearning,sun2025archhierarchicalhybridlearning,ahn2022icanisay,hu2023lookleapunveilingpower}. Intuitively, if a robot possesses reliable atomic skills and can seamlessly transition between them, any complex long-horizon task can be naturally resolved. This route relies entirely on atomic-only training data, forcing the model to genuinely recombine skills rather than rely on memorized long-horizon trajectories.
Although conceptually appealing, the true combinatorial generalization capability of this pipeline has not been systematically analyzed. To rigorously evaluate this setting, we introduce \textbf{\BenchName}, a benchmark dedicated to compositional manipulation in the absence of composed training trajectories. By restricting training data to isolated atomic-task trajectories and evaluating on unseen ordered compositions, \BenchName specifically targets the hierarchical-subtask route.
This design aligns with established protocols for testing generalization in long-horizon and lifelong manipulation~\citep{mees2022calvinbenchmarklanguageconditionedpolicy,liu2023liberobenchmarkingknowledgetransfer}.
In simulation, models are trained on isolated trajectories from 50 atomic tasks and evaluated on \NumCompositions novel multi-stage compositions. For real-robot evaluation, models are trained on 16 atomic manipulation tasks and tested on 13 multi-stage compositional tasks.
The benchmark further provides fine-grained metrics such as ordered success rates and transition success to diagnose failures across subtask boundaries.

To expose where failures arise within this route, we propose \textbf{\emph{Hierarchical Subtask Execution} (\textsc{HSE})}, a diagnostic execution framework that exposes skill recombination through a shared modular interface across the planner, trigger, and policy. Unlike prior hierarchical subtask frameworks that use hierarchy primarily as a solution architecture for long horizon execution, \textsc{HSE} uses hierarchy as a measurement interface: the subtask sequence, switching rule, and atomic policy can be controlled separately to reveal where atomic only recombination fails. This separation localizes failures into subtask sequencing errors, boundary timing errors, and downstream execution failures from handoff states.

The \BenchName evaluation with \textsc{HSE} reveals that atomic VLA training alone does not ensure reliable skill chaining. The bottleneck is a mismatch between the canonical start states used for atomic training and the handoff states in composed execution, where each downstream subtask starts from the endpoint of the preceding subtask. We refer to this mismatch as the \emph{start state gap}, which explains why strong atomic skills do not automatically transfer to reliable ordered execution.
This diagnosis suggests that improving compositional execution does not necessarily require collecting full composed demonstrations. We therefore introduce \textbf{\emph{Pose Robust Task Initialization} (\PRTI)}, which trains atomic policies on a broader set of task initial poses that resemble handoff states. This keeps the data regime atomic only while improving pose robustness for downstream subtasks.

In summary, our contributions are as follows:
$\spadesuit$ \textbf{Diagnostic Framework.}
We introduce \emph{Hierarchical Subtask Execution} (\textsc{HSE}), a diagnostic execution framework for the hierarchical tasks to localize failures in atomic skill recombination. 
$\spadesuit$ \textbf{Benchmark.}
We introduce \BenchName, a benchmark for compositional manipulation in the absence of composed training trajectories, with fine-grained metrics such as ordered success and transition success to diagnose failures across subtask boundaries.
$\spadesuit$ \textbf{Identifying and Addressing the Start-state Gap.}
We identify a critical mismatch between these unfamiliar handoff states and the canonical start of atomic training, which fundamentally hinders atomic-to-compositional transfer.
To address this gap, we introduce \emph{Pose-Robust Task Initialization} (\PRTI) to broaden the coverage of atomic start states.

\section{Related Work}
\label{sec:related-work}

\noindent\textbf{VLA models and full-instruction long-horizon training.}
Vision-language-action models provide the foundation for language-conditioned manipulation, with systems such as RT-2~\citep{brohan2023rt2visionlanguageactionmodelstransfer}, OpenVLA~\citep{kim2024openvlaopensourcevisionlanguageactionmodel}, Octo~\citep{octomodelteam2024octoopensourcegeneralistrobot}, $\pi_0$~\citep{black2024pi0visionlanguageactionflowmodel}, $\pi_{0.5}$~\citep{intelligence2025pi05visionlanguageactionmodelopenworld}, and Gemini Robotics~\citep{geminiroboticsteam2025geminiroboticsbringingai} transferring vision-language representations into robot action prediction at increasing scale.
Recent work also improves the action interface through diffusion-style heads, autoregressive--diffusion hybrids, action tokenization, vector-quantized actions, and spatial, force-aware, part-centric, or action-reasoning representations~\citep{wen2025dexvlavisionlanguagemodelplugin,wen2025diffusionvlageneralizableinterpretablerobot,liu2025hybridvlacollaborativediffusionautoregression,pertsch2025fastefficientactiontokenization,wang2025vqvlaimprovingvisionlanguageactionmodels,qu2025spatialvlaexploringspatialrepresentations,yu2025forcevlaenhancingvlamodels,huang2024copageneralroboticmanipulation,lee2025molmoactactionreasoningmodels}. For long-horizon tasks, a natural extension is the full-instruction route: train or adapt a VLA on complete multi-stage demonstrations and condition it on a composed command. Hierarchical VLM--VLA controllers, unified long-horizon policies, adaptive reasoning, and visual chain-of-thought systems instantiate this route~\citep{shi2025hirobotopenendedinstruction,yang2025lohovlaunifiedvisionlanguageactionmodel,lin2025onetwovlaunifiedvisionlanguageactionmodel,zhao2025cotvlavisualchainofthoughtreasoning}. These methods are complementary to ours because composed demonstrations are valuable when the target sequence distribution is covered. Our focus is on the stricter case in which composed trajectory coverage is finite and the evaluated ordered compositions are unseen. The question is not whether full-instruction training can help cover chains, but whether it demonstrates recombination beyond the sequence identities it observes.

\noindent\textbf{Hierarchical subtask execution, benchmarks, and handoff states.} The complementary route is to train only atomic skills and execute a composed command by switching the policy across known subtasks. This connects to skill composition, hierarchical imitation, task schemas, subpolicy switching, language-grounded action selection, VLM-guided planning, and routing or mixture-of-experts policies for long-horizon manipulation~\citep{strudel2020learningcombineprimitiveskills,fox2019hierarchicalvariationalimitationlearning,chitnis2020efficientbimanualmanipulationusing,sun2025archhierarchicalhybridlearning,ahn2022icanisay,hu2023lookleapunveilingpower,yang2025guiding,erdogan2025plan,zhou2024isr,xu2025moseskillbyskillmixtureofexpertslearning,yang2025drivemoemixtureofexpertsvisionlanguageactionmodel,reuss2024efficientdiffusiontransformerpolicies}. Benchmarks such as CALVIN and LIBERO have made language-conditioned sequence execution and transfer measurable~\citep{mees2022calvinbenchmarklanguageconditionedpolicy,liu2023liberobenchmarkingknowledgetransfer}, but they do not by themselves isolate the data boundary studied here: training contains isolated atomic trajectories, evaluation contains unseen ordered compositions. Existing long-horizon manipulation benchmarks evaluate sequential execution, but they often report aggregate task success and do not explicitly isolate whether failures arise from subtask selection, boundary timing, or launching a known downstream skill from the state produced by a previous skill.

\section{RoboCombine Benchmark and HSE Diagnostics}
\label{sec:benchname}

\subsection{RoboCombine Benchmark}

To rigorously evaluate compositional manipulation without composed training trajectories, we introduce \textbf{\BenchName}. By building on existing RoboTwin~\cite{chen2025robotwin} atomic tasks, \BenchName keeps the primitive-skill vocabulary familiar and shifts the evaluation focus to whether models can recombine already trained atomic skills into unseen ordered compositions.

\noindent\textbf{Training.}
Training is restricted to atomic task trajectories from RoboTwin, optionally with a VLA checkpoint trained on 50 atomic skills. No complete task sequence from the evaluation set is seen during training. Detailed \BenchName tasks and training settings are provided in Sec.~\ref{sec:RoboCombine_Details}.

\noindent\textbf{Testing.}
Each episode is a two-stage composition specified by an ordered pair of familiar atomic instructions $(s_1,s_2)$ and a long instruction ``$s_1$, then $s_2$''.
The evaluation records whether a model completes the composition in order and converts across the handoff boundary.

\noindent\textbf{Metrics.}
The primary metric is ordered success ($\mathrm{SR}_{\mathrm{ord}}$), which counts episodes in which both subtasks are completed in the specified order. We also report first- and second-subtask success ($\mathrm{SR}_{1}$ and $\mathrm{SR}_{2}$) and transition conversion ($\mathrm{Conv}_{1\rightarrow2}$), defined as the ordered-success rate among episodes that complete the first subtask.

\begin{figure}[t]
    \centering
    \includegraphics[width=\linewidth]{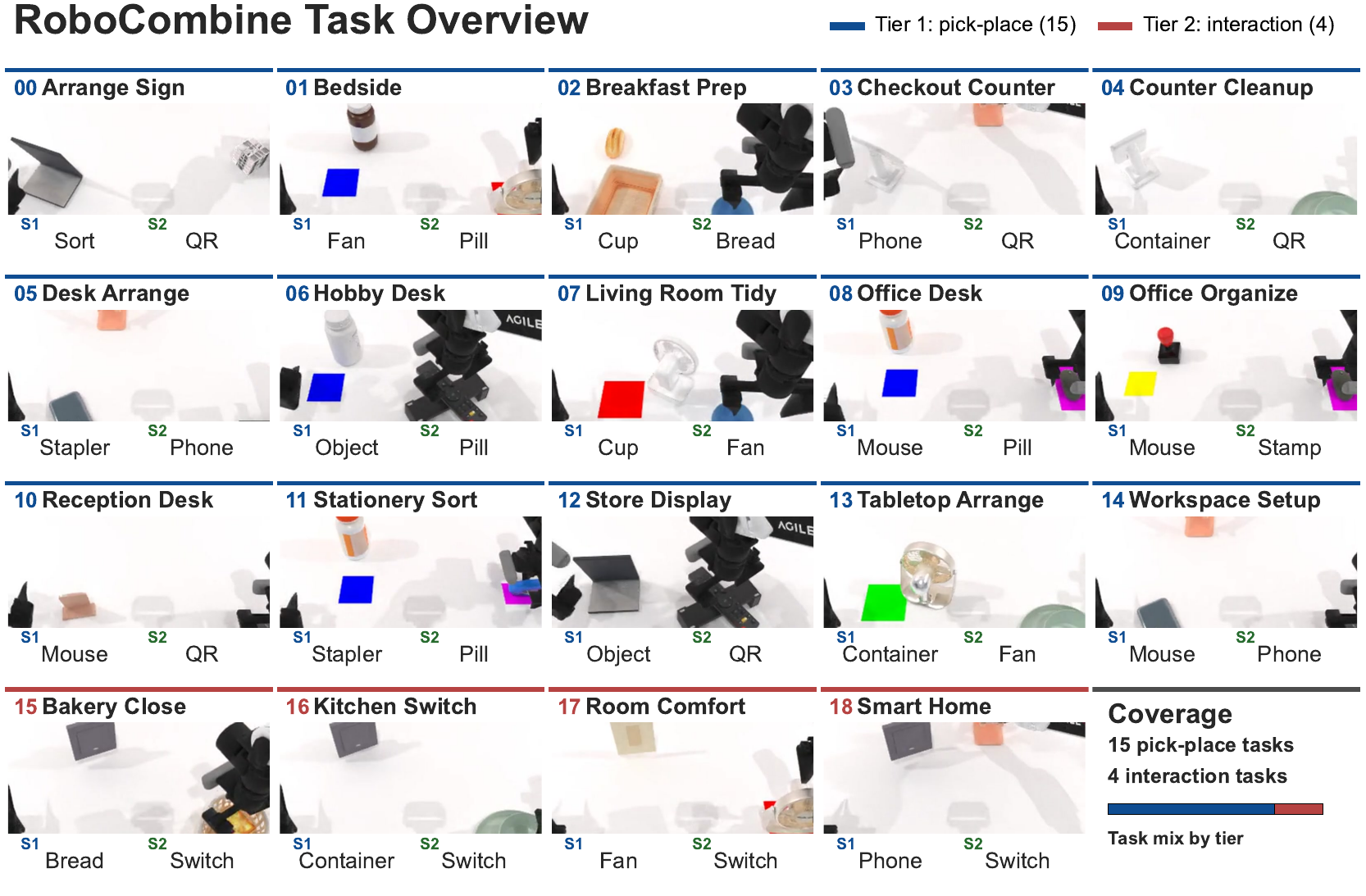}
    \vspace{-10pt}
    \caption{\textbf{RoboCombine task overview.}
    Each row shows one composition task.
    The target subtask starts from the source terminal state; full instructions are listed in Tab.~\ref{tab:robocombine_task_instructions}.}
    \label{fig:core19_tasks_overview}
    \vspace{-10pt}
\end{figure}

Fig.~\ref{fig:core19_tasks_overview} visualizes the ordered compositions and the handoff that each task induces. The benchmark contains 20 evaluation rollouts per composition, for a total of 380 episodes per policy variant. These compositions form two scenario families: 15 \emph{pick and place} or object-arrangement compositions and four \emph{placement to interaction} compositions.

\subsection{HSE Diagnostic Framework}
\label{sec:hase}
\label{sec:evaluation_tracks}

\begin{figure*}[!t]
    \centering
    \includegraphics[width=0.98\textwidth]{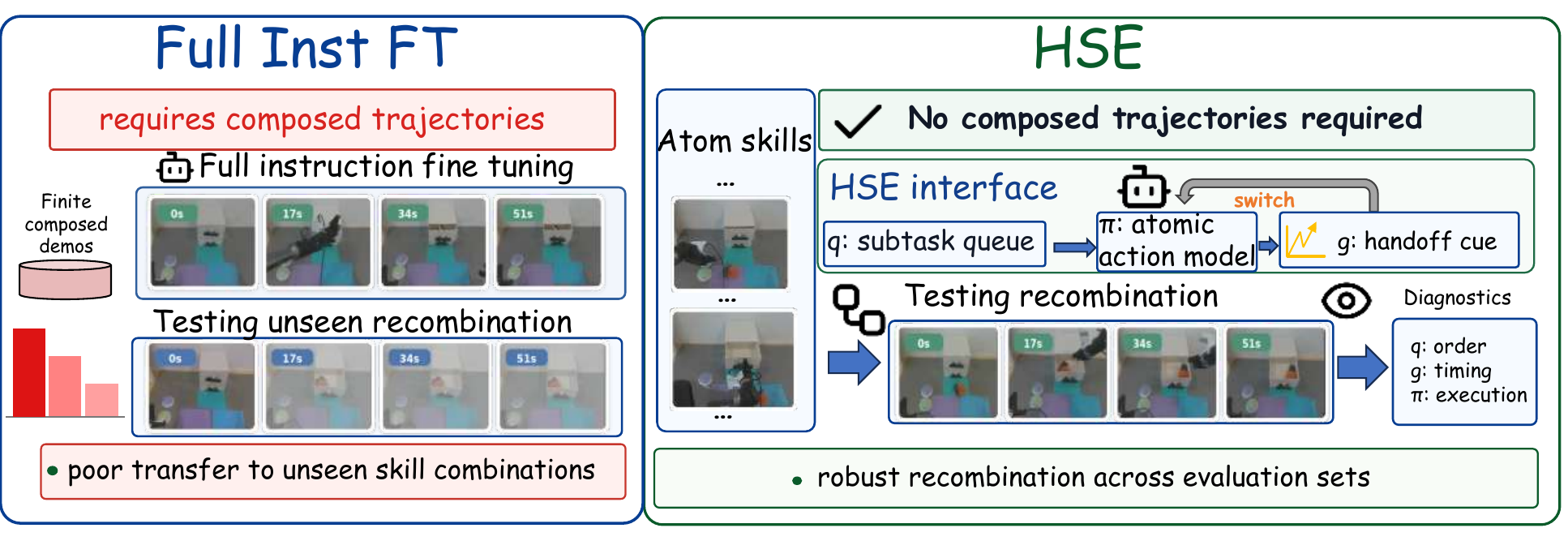}
    \vspace{-5pt}
    \caption{\textbf{Full instruction fine tuning and \textsc{HSE} make different recombination assumptions.}
For a command such as ``open the drawer, then place the persimmon from the right plate into the drawer, then close the drawer,''
Full Inst FT treats it as one long horizon instruction supervised by composed trajectories, which can transfer poorly to unseen skill combinations.
\textsc{HSE} instead decomposes it into a subtask queue $q$, switches by a handoff cue $g$, and invokes the atomic policy $\pi$ one subtask at a time, exposing order, timing, and execution failures.}
    \label{fig:fullinst_vs_hse_handdrawn}
    \vspace{-10pt}
\end{figure*}

Aggregate success rates provide limited diagnostic information for composed manipulation. They do not distinguish failures caused by command interpretation, subtask ordering, handoff timing, or downstream execution from those caused by handoff states. 
\textsc{HSE} is our diagnostic instantiation of the hierarchical subtask route.
It is model agnostic: it treats the underlying VLA as an instruction-conditioned atomic executor and only controls which atomic instruction is active and when that instruction is replaced.
It executes a composed command by keeping one atomic instruction active at a time and replacing that instruction when a handoff cue fires.
This gives three controllable factors: \emph{the subtask queue, the handoff cue, and the atomic VLA model}.
The queue determines what should be executed next, the cue determines when control should move to the next subtask, and the model determines whether the active atomic instruction can be executed from the current physical state.
By fixing or replacing these factors across matched execution tracks, \textsc{HSE} gives each performance gap a concrete diagnostic meaning.
Fig.~\ref{fig:fullinst_vs_hse_handdrawn} contrasts this modular \textsc{HSE} interface with full-instruction fine-tuning, which instead treats a composed command as a single long-horizon imitation target.

Let $\mathcal{Z}=\{\task{1},\ldots,\task{K}\}$ be the atomic task vocabulary during training.
A VLA model takes an observation history $h_t=o_{t-H+1:t}$ and an active instruction $\ell\in\mathcal{Z}$, and predicts an action chunk:
\begin{equation}
    a_{t:t+L-1} \sim \pi_\theta(\cdot \mid h_t,\ell).
\end{equation}

Given a composed command $c$, a decomposition module produces a queue of atomic subtasks:
\begin{equation}
    q=\mathcal{D}(c)=(q_1,\ldots,q_m),
    \qquad q_j\in\mathcal{Z}.
\end{equation}

\textsc{HSE} executes this queue at the level of model invocations.
Let $\tau_k$ denote the time of the $k$th policy invocation, and let $j_k$ denote the active subtask index for that invocation.
At invocation $k$, the robot executes only the current atomic instruction $q_{j_k}$:
\begin{equation}
    a_{\tau_k:\tau_k+L-1}
    \sim
    \pi(\cdot \mid h_{\tau_k}, q_{j_k}).
\end{equation}

The handoff cue is monitored during this chunk's execution.
If the cue fires at any time within the current chunk, it is latched until the next policy invocation:
\begin{equation}
    G_k=\max_{\tau_k \le s < \tau_k+L} g_s,
    \qquad g_s\in\{0,1\}.
\end{equation}
The active instruction is then updated only before the next chunk is queried:
\begin{equation}
    j_1=1,\qquad
    j_{k+1}=\min\{j_k+G_k,m\}.
\end{equation}
Thus, a handoff cue does not change the instruction inside an already predicted action chunk.
It changes which atomic instruction conditions the next model invocation.
This execution can be summarized as a tuple $(q,g,\pi)$, where $q$ specifies the subtask order, $g$ specifies the handoff rule that produces the latched cue $G_k$, and $\pi$ is the model used at each stage.

The diagnostic role of \textsc{HSE} comes from evaluating matched modes that differ in only one factor:
\begin{equation}
\begin{array}{ll}
\mathrm{Pred}:     (\hat q,\hat g,\pi_\theta),      &
\mathrm{Ordered}:  (q^\star,\hat g,\pi_\theta), \\
\mathrm{Oracle}:   (q^\star,g^\star,\pi_\theta),    &
\mathrm{Enhanced}: (q^\star,g^\star,\pi_{\theta^+}).
\end{array}
\end{equation}
Here \emph{Pred} uses the predicted queue and predicted cue.
\emph{Ordered} fixes the subtask queue to $q^\star$.
\emph{Oracle} further fixes the handoff cue to $g^\star$.
\emph{Enhanced} uses the same oracle queue and cue as \emph{Oracle}, but replaces the atomic policy with $\pi_{\theta^+}$.
Implementation details of the task decomposition module and the progress predictor used to instantiate $\hat q$ and $\hat g$ are provided in Appendix~\ref{sec:HSE_Details}.

Let $Y(M)$ denote ordered success under \textsc{HSE} mode $M$.
The gap between $Y(\mathrm{Pred})$ and $Y(\mathrm{Ordered})$ measures the effect of fixing the subtask queue.
The gap between $Y(\mathrm{Ordered})$ and $Y(\mathrm{Oracle})$ measures the effect of fixing the handoff cue.
The gap between $Y(\mathrm{Oracle})$ and $Y(\mathrm{Enhanced})$ measures the effect of replacing the atomic model while keeping the same queue and cue.
Thus, \textsc{HSE} is diagnostic because each performance gap corresponds to a controlled change in the queue, cue, or model, rather than to an unspecified change in the whole composed task. 

\section{Analysis and Pose Robust Repair}
\label{sec:method}

\subsection{Diagnostic Analysis}
\label{sec:problem}
\label{sec:start_state_gap}
\noindent\textbf{Full Instruction Fine Tuning across Evaluation Sets.}
Our evaluation of the direct \textsc{Full Inst FT} route reveals a strong dependence on the composition set used during training.
We start from a VLA model trained on isolated atomic task trajectories, continue training it on a mixture of isolated atomic trajectories and composed trajectories, and assess ordered success on the three composition sets shown in Fig.~\ref{fig:intro_combo_ckpt_trend}.
The results are telling.
The model achieves 40.0\% ordered success on set A, but drops to 15.0\% on set B and 4.4\% on set C, with only 7.7\% averaged over sets B and C.
This finding establishes the first diagnostic principle: finite full instruction supervision improves the seen compositions but provides limited evidence of robust skill recombination on other evaluation sets.

\begin{table}[!t]
\centering
\caption{\textbf{Per-composition ordered success grouped by policy and execution mode.}
Each row is one RoboCombine composition and reports ordered success (\%) over 20 rollouts.
Under each policy, the three subcolumns correspond to \textsc{Oracle-HSE}, \textsc{Ordered-HSE}, and \textsc{Full-Inst}.
\textsc{Oracle-HSE} denotes the oracle \textsc{HSE} track with ground-truth subtask order and handoff cue,
and \textsc{Ordered-HSE} denotes the ordered \textsc{HSE} track with ground-truth subtask order and predicted handoff cue.
Aggregate ordered success rates ($n=380$ per policy variant) are reported in the final row.}
\label{tab:core19_per_task_policy_mode}
\begingroup
\scriptsize
\renewcommand{\arraystretch}{1.08}
\setlength{\tabcolsep}{3.2pt}
\newcommand{\zero}{\textcolor{black!28}{0}}
\resizebox{\textwidth}{!}{\begin{tabular}{@{}lcccccccccccc@{}}
\toprule
Task
& \multicolumn{3}{c}{Motus~\cite{bi2025motus}}
& \multicolumn{3}{c}{FastWAM~\cite{yuan2026fastwam}}
& \multicolumn{3}{c}{LingBot-VLA~\cite{wu2026pragmatic}}
& \multicolumn{3}{c}{$\pi_{0.5}$~\cite{intelligence2025pi05visionlanguageactionmodelopenworld}} \\
\cmidrule(lr){2-4}\cmidrule(lr){5-7}\cmidrule(lr){8-10}\cmidrule(lr){11-13}
& Oracle & Ordered & Full Inst
& Oracle & Ordered & Full Inst
& Oracle & Ordered & Full Inst
& Oracle & Ordered & Full Inst \\
\midrule
Arrange Sign       & 10.0 & 5.0  & \zero & \zero & \zero & \zero & \zero & \zero & \zero & \zero & \zero & \zero \\
Bedside            & 30.0 & 25.0 & 30.0  & \zero & \zero & \zero & \zero & \zero & 15.0  & 5.0   & 5.0   & \zero \\
Breakfast Prep     & 75.0 & 80.0 & 20.0  & 10.0  & 5.0   & 5.0   & \zero & \zero & \zero & 10.0  & \zero & 5.0 \\
Checkout Counter   & 30.0 & 15.0 & 15.0  & \zero & \zero & 5.0   & \zero & 5.0   & \zero & 5.0   & \zero & \zero \\
Counter Cleanup    & 15.0 & 10.0 & 15.0  & \zero & \zero & \zero & \zero & \zero & \zero & \zero & \zero & \zero \\
Desk Arrange       & 10.0 & 10.0 & 5.0   & 5.0   & 5.0   & \zero & \zero & \zero & \zero & \zero & \zero & \zero \\
Hobby Desk         & 35.0 & 10.0 & 15.0  & 10.0  & 10.0  & \zero & 40.0  & 40.0  & 20.0  & \zero & \zero & \zero \\
Living Room Tidy   & 10.0 & 30.0 & 5.0   & 75.0  & 20.0  & 35.0  & 5.0   & 5.0   & 5.0   & 5.0   & \zero & \zero \\
Office Desk        & 15.0 & 35.0 & 10.0  & \zero & \zero & \zero & 5.0   & \zero & 5.0   & \zero & \zero & \zero \\
Office Organize    & 15.0 & 25.0 & 15.0  & \zero & \zero & \zero & 5.0   & \zero & 5.0   & \zero & \zero & \zero \\
Reception Desk     & 15.0 & 20.0 & 5.0   & \zero & \zero & 5.0   & \zero & \zero & \zero & \zero & \zero & \zero \\
Stationery Sort    & 10.0 & 5.0  & 10.0  & \zero & 5.0   & \zero & 5.0   & \zero & \zero & \zero & \zero & \zero \\
Store Display      & 20.0 & 20.0 & 5.0   & 5.0   & 15.0  & \zero & \zero & \zero & \zero & \zero & \zero & \zero \\
Tabletop Arrange   & \zero& \zero& 20.0  & \zero & \zero & \zero & \zero & \zero & \zero & \zero & \zero & \zero \\
Workspace Setup    & 50.0 & 25.0 & 15.0  & \zero & \zero & \zero & \zero & \zero & \zero & \zero & \zero & \zero \\
Bakery Close       & 35.0 & 10.0 & 5.0   & 10.0  & 10.0  & \zero & \zero & \zero & \zero & \zero & \zero & \zero \\
Kitchen Switch     & \zero& \zero& \zero & \zero & \zero & \zero & \zero & \zero & \zero & 5.0   & \zero & \zero \\
Room Comfort       & \zero& \zero& \zero & \zero & \zero & \zero & \zero & \zero & \zero & \zero & \zero & \zero \\
Smart Home         & \zero& 5.0  & \zero & \zero & \zero & \zero & \zero & \zero & \zero & \zero & 10.0  & \zero \\
\midrule
Overall SR
& \textbf{19.7} & 17.4 & 10.0
& \textbf{6.1} & 3.7 & 2.6
& \textbf{3.2} & 2.6 & 2.6
& \textbf{1.6} & 0.8 & 0.3 \\
\bottomrule
\end{tabular}}
\endgroup
\end{table}

\begin{figure}[!t]
    \centering
    \includegraphics[width=0.98\linewidth]{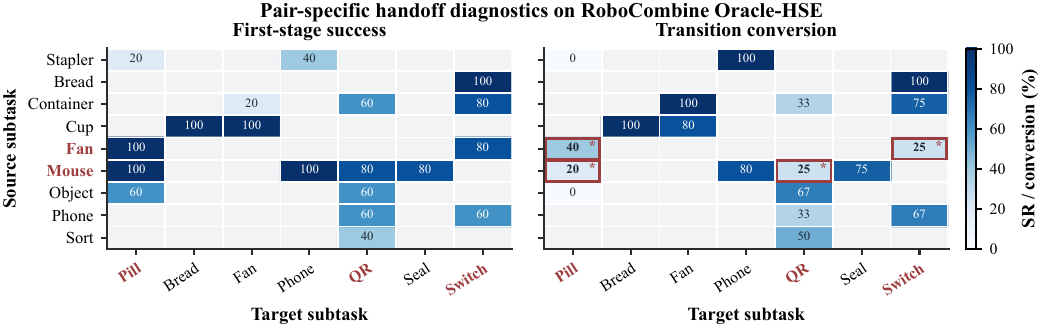}
    \caption{\textbf{RoboCombine pairwise handoff diagnostics under oracle timed \textsc{HSE}.}
We evaluate Motus on \BenchName under the oracle \textsc{HSE} track, which supplies the ground-truth subtask order and handoff cue. Each cell corresponds to one ordered composition, with color and number indicating the metric value.
The left matrix reports first-stage success for the source subtask, and the right matrix reports transition conversion, i.e., the success rate of the target subtask conditioned on successful completion of the source subtask.}
    \label{fig:transition_matrix}
\end{figure}

\noindent\textbf{Atomic Skill Recombination under \textsc{HSE}.}
We next evaluate atomic skill recombination under the oracle timed \textsc{HSE} track.
This diagnostic track supplies both the ground truth subtask order and the handoff cue.
The evaluation instead asks a narrower execution question:
Given the correct active atomic instruction, can the model execute it from the physical state left by the preceding subtask?
Tab.~\ref{tab:core19_per_task_policy_mode} shows that performance under this controlled interface varies substantially across models over 20 rollouts per composition.
Motus attains the highest aggregate ordered success in all three execution modes: 19.7\% with Oracle-HSE, 17.4\% with Ordered-HSE, and 10.0\% with Full-Inst. The difference between its two HSE modes is not statistically distinguishable in a task-paired comparison ($p\approx0.44$).
This variation shows that some atomic executors can use the supplied subtask order and handoff cue to complete composed tasks, but that the correct interface alone is not sufficient.
For models that remain near failure, the bottleneck lies in the executor itself:
the downstream atomic skill must start from a handoff state rather than from the canonical start distribution seen during isolated atomic training.
Aggregate success does not reveal which handoffs cause this failure.
We therefore inspect the same oracle \textsc{HSE} setting at the source target level.
Fig.~\ref{fig:transition_matrix} shows that first stage completion does not reliably convert into downstream execution.
Several source subtasks can be completed successfully, yet their transition conversion remains much lower.
Moreover, conversion is pair specific:
the same downstream subtask can succeed or fail depending on which source subtask precedes it, and the same source subtask can lead to different conversion rates across targets.
This pattern indicates that the bottleneck is not simply whether an atomic skill exists, but whether it can launch from the particular handoff state produced by the preceding skill.
\FloatBarrier

\subsection{Pose Robust Task Initialization}
\label{sec:prti}
The diagnosis in Sec.~\ref{sec:problem} identifies a major bottleneck in ordered execution. Downstream atomic skills are often invoked from handoff states rather than from the canonical starts seen during atomic training. We therefore introduce Pose Robust Task Initialization (\PRTI), an atomic post-training strategy that broadens the start-state coverage of each skill without using composed demonstrations. \PRTI builds an auxiliary pool $\mathcal{S}$ from late segments of successful atomic trajectories, capturing object poses, end effector poses, and contact configurations likely to appear near subtask boundaries. For each target skill $\task{i}$, we use the task valid subset $\mathcal{S}_i\subseteq\mathcal{S}$, containing states from which $\task{i}$ remains executable.

In simulation, a data engine restarts single skill trajectories from states sampled from $\mathcal{S}_i$. On hardware, we can collect additional single skill demonstrations from randomized, task valid initial configurations by perturbing relevant object poses, robot approach poses, and fixture states. For the experiments reported in this paper, \PRTI initial states are sampled only from pre-existing successful atomic demonstrations.

During post training, we mix canonical atomic trajectories and \PRTI trajectories at a 1:1 ratio. The policy is always conditioned only on the target atomic instruction $\task{i}$, so \PRTI expands atomic start state coverage while preserving the atomic only training boundary.

The resulting objective is
\begin{equation}
\label{eq:prti}
    \min_{\theta}
    \mathbb{E}_{i \sim p(\mathcal{Z}),\,
    x_0 \sim m_i^\alpha}
    \left[
    \mathcal{L}_{\mathrm{atomic}}(\theta;x_0,\task{i})
    \right],
\end{equation}
where $m_i^\alpha$ mixes the canonical start distribution $p_i^{\mathrm{atom}}$ and the task valid auxiliary start distribution $\tilde{q}_i$ with noncanonical start fraction $\alpha$.

\section{Experiments}
We conduct experiments to address the questions below:
\textbf{Q1: Pose Robust Task Initialization (Sec.~\ref{Q1}).}
Does \PRTI broaden atomic start state support and improve downstream execution while preserving the atomic only data boundary?
\textbf{Q2: Real World Deployment (Sec.~\ref{Q2}).}
Can the same HSE interface run variable length real robot sequences under variable scenes, and does \PRTI improve execution across multiple physical handoffs?

\subsection{Experimental Setup}
\label{baseline}

All simulation experiments use \BenchName from Sec.~\ref{sec:benchname}.
For diagnostic \textsc{HSE} evaluations, we use the oracle HSE.
This setting removes subtask-ordering and boundary-timing uncertainty, so remaining failures measure whether the atomic policy can execute the active downstream instruction from the handoff state produced by the preceding subtask.
Real-robot experiments use the \emph{Order} \textsc{HSE} mode, where the subtask queue is fixed to the specified order $q^\star$ and the progress predictor triggers handoffs online, without benchmark-supplied boundary cues.
We report ordered success ($\mathrm{SR}_{\mathrm{ord}}$), first and second subtask success ($\mathrm{SR}_{1}$ and $\mathrm{SR}_{2}$), and transition conversion ($\mathrm{Conv}_{1\rightarrow2}$), following Sec.~\ref{sec:benchname}. Detailed \BenchName tasks and training settings are provided in Sec.~\ref{sec:RoboCombine_Details}.
\subsection{\PRTI Broadens Atomic Start State Support}
\label{Q1}

We evaluate \PRTI under the oracle \textsc{HSE} setting, where the benchmark supplies both the subtask queue $q^\star$ and the handoff cue $g^\star$.
This removes ordering and boundary-timing uncertainty, so differences in ordered success directly measure whether the invoked atomic model can execute downstream subtasks from handoff states.
We report the base atomic policy as a reference and compare \PRTI with two complementary controls: unstructured random initial-state perturbations and an execution-time reset oracle that returns the robot to its canonical pose between subtasks.

\begin{table*}[t]
\centering
\caption{\textbf{Ordered success across compositions.} Each task uses 20 episodes per method. All values are percentages. Only tasks with nonzero success under at least one matched HSE execution are shown. Summary metrics use all 19 tasks.
Random uses unstructured perturbations of initial states as a control for generic state diversity.
Reset restores only the robot's canonical pose between subtasks, leaving the preceding subtask's object and contact states unchanged.}
\label{tab:prti_reset_controls}
\begingroup
\fontsize{7.5}{9}\selectfont
\setlength{\tabcolsep}{2.6pt}
\renewcommand{\arraystretch}{1.10}
\begin{minipage}{0.88\textwidth}
\centering
\begin{tabular*}{\linewidth}{@{\extracolsep{\fill}}lrrrr@{\hspace{12pt}}lrrrr@{}}
\toprule
Task & Base & Random & Reset & PRTI & Task & Base & Random & Reset & PRTI \\
\midrule
Arrange Sign & 5 & \textcolor{black!28}{0} & \textbf{10} & 5 & Office Organize & 15 & \textcolor{black!28}{0} & 5 & \textbf{20} \\
Bedside & \textcolor{black!28}{0} & 5 & 10 & \textbf{25} & Reception Desk & 5 & \textcolor{black!28}{0} & 10 & \textbf{15} \\
Breakfast Prep & 5 & 20 & \textbf{40} & 30 & Stationery Sort & \textcolor{black!28}{0} & 5 & 10 & \textbf{15} \\
Checkout Counter & \textbf{5} & \textcolor{black!28}{0} & \textcolor{black!28}{0} & \textcolor{black!28}{0} & Store Display & \textcolor{black!28}{0} & \textcolor{black!28}{0} & \textcolor{black!28}{0} & \textbf{20} \\
Counter Cleanup & \textcolor{black!28}{0} & \textcolor{black!28}{0} & 5 & \textbf{35} & Tabletop Arrange & 5 & \textcolor{black!28}{0} & \textcolor{black!28}{0} & \textbf{15} \\
Desk Arrange & \textcolor{black!28}{0} & 10 & 10 & \textbf{20} & Workspace Setup & 5 & 20 & 15 & \textbf{30} \\
Hobby Desk & 5 & 10 & 15 & \textbf{20} & Bakery Close & \textcolor{black!28}{0} & 5 & 10 & \textbf{15} \\
Living Room Tidy & \textcolor{black!28}{0} & \textcolor{black!28}{0} & 25 & \textbf{40} & Kitchen Switch & \textcolor{black!28}{0} & \textcolor{black!28}{0} & 10 & \textbf{20} \\
Office Desk & \textcolor{black!28}{0} & \textbf{5} & \textbf{5} & \textbf{5} & Room Comfort & \textcolor{black!28}{0} & \textcolor{black!28}{0} & \textcolor{black!28}{0} & \textbf{15} \\
\end{tabular*}
\par\vspace{3pt}
\begin{tabular*}{\linewidth}{@{\extracolsep{\fill}}p{0.48\linewidth}rrrr@{}}
\midrule
\textbf{All 19 tasks} & Base & Random & Reset & PRTI \\
\addlinespace[2pt]
Nonzero-task coverage & 42.11 & 42.11 & 73.68 & \textbf{89.47} \\
Avg. ordered success & 2.63 & 4.21 & 9.47 & \textbf{18.16} \\
Transition conversion $\mathrm{Conv}_{1\rightarrow2}$ & 7.52 & 12.50 & 29.75 & \textbf{39.88} \\
\bottomrule
\end{tabular*}
\end{minipage}
\endgroup
\end{table*}

Tab.~\ref{tab:prti_reset_controls} reports an ordered success rate of 2.63\% for the base atomic policy. Random perturbation achieves 4.21\%, the reset oracle achieves 9.47\%, and \PRTI achieves 18.16\%.
The 13.9\% absolute gain over random perturbation shows that unstructured state diversity alone is insufficient; trajectory-supported, physically feasible initial states provide substantially stronger handoff robustness.
Resetting only the robot recovers part of the gap but does not restore the object and contact states produced by the preceding subtask, whereas \PRTI executes directly from the complete handoff state without an artificial reset.
\subsection{Real-World Deployment}\label{Q2}
For \textbf{Q2}, the real-world experiments test whether the simulation conclusion transfers to physical execution: subtask-conditioned HSE remains more reliable than the finite \textsc{Full-Inst-FT} route when both are deployed on the same robot scenes.

\newcolumntype{C}[1]{>{\centering\arraybackslash}p{#1}}

\begin{table*}[t]
\centering
\caption{Real-robot mean progress on compositional tasks. Each entry averages five physical rollouts using the task-length-specific ordered-milestone rubric in Sec.~\ref{sec:real_robot_mean_progress}. Each task ID denotes an ordered sequence of atomic tasks; detailed definitions of the atomic task IDs are provided in Sec.~\ref{sec:real_robot_deatils}. \textsc{Full-Inst-FT} is trained with the collected Set A compositions. \textsc{Oracle-HSE+PRTI} differs from \textsc{Oracle-HSE} only by PRTI atomic post-training; neither HSE variant uses composed trajectories.}
\vspace{5pt}
\label{tab:real_robot_results}
\small
\setlength{\tabcolsep}{5pt}
\renewcommand{\arraystretch}{1.18}

\resizebox{\textwidth}{!}{\begin{tabular}{@{}C{0.18\textwidth}C{0.12\textwidth}!{\color{gray!45}\vrule width 0.5pt}C{0.12\textwidth}C{0.13\textwidth}C{0.12\textwidth}C{0.12\textwidth}C{0.13\textwidth}@{}}
\toprule
\rowcolor{gray!12}
\textbf{Task Split} &
\textbf{Task ID} &
\textbf{FULL-INST} &
\textbf{FULL-INST-FT} &
\textbf{Order-HSE} &
\textbf{Oracle-HSE} &
\shortstack{\textbf{Oracle-HSE}\\\textbf{+ PRTI}} \\
\midrule

\multirow{4}{*}{\shortstack[c]{A: Collected\\Compositions}}
& 1-7           & 0.62 & \textbf{0.94} & 0.82 & 0.76 & 0.82 \\
& 1-8-6         & 0.24 & 0.62 & 0.34 & 0.44 & \textbf{0.66} \\
& 1-15-11-16    & 0.08 & \textbf{0.64} & 0.46 & 0.38 & 0.52 \\
& 7-5-8-6-10    & 0.06 & 0.06 & 0.46 & 0.62 & \textbf{0.72} \\
\midrule

\multirow{5}{*}{\shortstack[c]{B: Overlap\\Held-Out}}
& 1-12          & 0.04 & 0.88 & 0.88 & \textbf{0.94} & \textbf{0.94} \\
& 1-2           & 0.20 & 0.50 & 0.94 & \textbf{1.00} & \textbf{1.00} \\
& 1-6-2         & 0.00 & 0.42 & 0.44 & \textbf{0.52} & 0.44 \\
& 1-5-2-10      & 0.00 & 0.24 & 0.64 & 0.82 & \textbf{0.88} \\
& 4-1-6-2-10    & 0.16 & 0.02 & 0.66 & 0.70 & \textbf{0.74} \\
\midrule

\multirow{4}{*}{\shortstack[c]{C: Disjoint\\Held-Out}}
& 3-10          & 0.20 & 0.00 & \textbf{0.94} & 0.88 & \textbf{0.94} \\
& 4-10          & 0.38 & 0.08 & \textbf{0.94} & \textbf{0.94} & \textbf{0.94} \\
& 9-3-10        & 0.10 & 0.06 & 0.72 & \textbf{0.94} & \textbf{0.94} \\
& 9-14-10       & 0.02 & 0.06 & 0.48 & 0.80 & \textbf{0.88} \\
\midrule

\rowcolor{gray!10}
\multicolumn{2}{c!{\color{gray!45}\vrule width 0.5pt}}{\textbf{Average}}
& 0.16 & 0.35 & 0.67 & 0.75 & \textbf{0.80} \\
\bottomrule
\end{tabular}}
\end{table*}

\noindent\textbf{Tasks and training details.}
We build our real-robot benchmark on a dual-arm AgileX robotic platform equipped with three camera views. The benchmark contains 16 atomic tasks spanning object-interaction skills, including button pressing and drawer opening/closing, as well as pick-and-place manipulation skills. Starting from the $\pi_{0.5}$, we fine-tune the model on these atomic tasks using NVIDIA A800 GPUs, and deploy the policy for real-robot inference on an NVIDIA RTX 4090 GPU. We optimize the model with AdamW, apply gradient-norm clipping with a threshold of 1.0, and maintain an exponential moving average of model parameters with a decay rate of 0.999. To evaluate compositional generalization beyond the atomic training distribution, we design 13 compositional tasks, each composed of two to five atomic steps, and use them as our real-robot evaluation suite. \PRTI is evaluated on the same hardware, task instances, execution protocol, scoring rubric, and five-rollout budget as \textsc{Oracle-HSE}, with \PRTI as the only added intervention. Further details on the real-robot task instructions, atomic-task success rates, and mean-progress metric are provided in Sec.~\ref{sec:real_robot_deatils}.

\begin{figure*}[!t]
    \centering
    \includegraphics[width=0.9\linewidth]{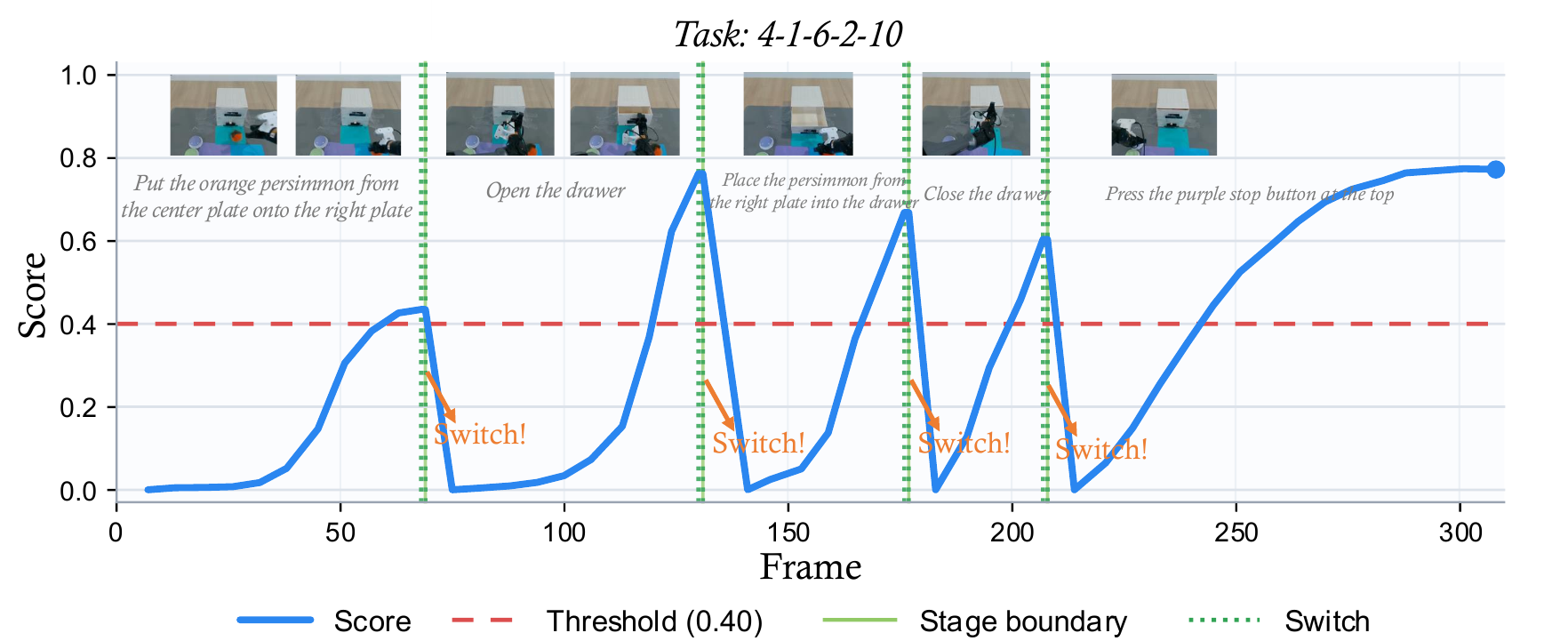}
    \caption{\textbf{Real-robot execution trace on the five-step compositional task \textit{4-1-6-2-10}}. The top row shows representative frames from the rollout, and the blue curve plots the progress predictor score over time. The score rises within each stage and drops after a successful switch, illustrating hierarchical execution over the full task sequence.}
    \vspace{-15pt}
    \label{fig:5task_vis}
\end{figure*}

\noindent\textbf{Real-robot compositional generalization.}
We evaluate 13 real-robot compositional tasks consisting of variable-length 2--5 step sequences. The tasks are divided into three splits: Set A contains seen compositions, Set B contains held-out compositions with partial overlap to Set A, and Set C contains unseen compositions. This setup evaluates whether a method can generalize beyond memorized task compositions. Tab.~\ref{tab:real_robot_results} reports mean progress under a shared ordered-milestone rubric and shows that our hierarchical subtask execution framework consistently outperforms direct full-instruction execution. Across all tasks, \textsc{Order-HSE} achieves an average mean progress of $0.67$, compared with $0.16$ for \textsc{Full-Inst} and $0.35$ for \textsc{Full-Inst-FT}. While \textsc{Full-Inst-FT} performs well on seen compositions, its performance drops significantly on unseen Set C tasks. In contrast, \textsc{Order-HSE} maintains strong performance on both held-out and unseen compositions, achieving $0.71$ on Set B and $0.77$ on Set C. Under the matched oracle switching protocol, adding \PRTI increases average real-robot mean progress from $0.75$ to $0.80$. \PRTI improves 8 of the 13 tasks, matches \textsc{Oracle-HSE} on four tasks, and decreases on one task, with improvements spanning 2--5-step compositions and therefore multiple handoff boundaries. These results indicate that fine-tuning on full compositional instructions mainly benefits familiar compositions, whereas subtask-level execution and broader atomic start-state support better support compositional generalization. Visualizations of compositional task execution are provided in Fig.~\ref{fig:5task_vis} and Appendix~\ref{sec:real_robot_deatils}.
By reusing atomic skills and switching between them online, our framework avoids learning each long-horizon composition separately and transfers more robustly to novel multi-step real-robot tasks.

\section{Conclusion}
\label{sec:conclusion}
We introduce \BenchName, a benchmark for atomic-only skill recombination, and \textsc{HSE}, a diagnostic protocol for full-instruction execution and oracle/progress-based handoffs.
Our results show that
(1) finite full-instruction training mainly fits seen compositions instead of recombining skills in unseen ones;
(2) even with correct subtask order and handoff cues, downstream subtasks remain unreliable when launched from handoff rather than canonical atomic states;
and (3) \PRTI mitigates this gap by broadening atomic start-state support, improving ordered execution without composed training trajectories.

\subsection*{AI Use Statement}
We use a large language model (LLM) as the task-decomposition module in HSE.
Given a composed language instruction and the closed atomic-task vocabulary,
the LLM rewrites the instruction as an ordered list of valid atomic subtasks.
We constrain the output to atomic-task labels and parse away any additional
text. The LLM is used only as a language interface for selecting and ordering
atomic skills; it does not generate robot actions or replace the VLA policy.
The authors reviewed the LLM-assisted component and take responsibility for
the methodology, experiments, and conclusions presented in this work.

\bibliography{refs}
\bibliographystyle{plainnat}

\clearpage
\appendix

\section{Limitations and future work}
\label{sec:Limitations}

\noindent\textbf{Limitations.}
This work studies compositional manipulation under an atomic only data regime, where evaluated compositions are built from familiar atomic task labels. Thus, our conclusions should be interpreted as evidence about recombining known skills, not as generalization to unseen atomic tasks, new embodiments, or open ended real world environments. 

\noindent\textbf{Future Work.}
Future work can extend \BenchName and \textsc{HSE} to longer and more diverse task chains, including variable length compositions and repeated atomic subtasks. The same diagnostic framework can also be used to evaluate how VLA models improve under handoff aware training, richer progress supervision, and broader atomic start state coverage. Another important direction is to move beyond controlled handoff analysis toward fully autonomous systems, where decomposition, progress prediction, recovery, and execution are learned and evaluated jointly in both simulation and real world manipulation settings.

\section{RoboCombine Details}
\label{sec:RoboCombine_Details}
This appendix provides the complete task specification and atomic skill context for \BenchName.
Table~\ref{tab:robocombine_task_instructions} lists the evaluated RoboCombine compositions and their corresponding atomic instructions, while Table~\ref{tab:atomic_clean_randomized_success} reports the isolated RoboTwin atomic success rates used to contextualize each model's base skill competence before composition.
\begin{table}[htbp]
\caption{\textbf{Full RoboCombine task instructions.}
The main overview figure uses short labels for readability; this table gives the complete source and target instructions for each ordered composition.}
\label{tab:robocombine_task_instructions}
\centering
\setlength{\tabcolsep}{4.2pt}
\renewcommand{\arraystretch}{1.08}
\resizebox{0.99\linewidth}{!}{\begin{tabular}{c l l l l}
\toprule
ID & Scenario & Family & Source instruction $s_1$ & Target instruction $s_2$ \\
\midrule
00 & Arrange Sign & pick-place & Sort objects & Rotate QR code \\
01 & Bedside & pick-place & Place fan on pad & Move pill bottle to pad \\
02 & Breakfast Prep & pick-place & Place cup on coaster & Place bread in basket \\
03 & Checkout Counter & pick-place & Place phone on stand & Rotate QR code \\
04 & Counter Cleanup & pick-place & Place container on plate & Rotate QR code \\
05 & Desk Arrange & pick-place & Move stapler to pad & Place phone on stand \\
06 & Hobby Desk & pick-place & Place object on stand & Move pill bottle to pad \\
07 & Living Room Tidy & pick-place & Place cup on coaster & Place fan on pad \\
08 & Office Desk & pick-place & Place mouse on pad & Move pill bottle to pad \\
09 & Office Organize & pick-place & Place mouse on pad & Stamp seal \\
10 & Reception Desk & pick-place & Place mouse on pad & Rotate QR code \\
11 & Stationery Sort & pick-place & Move stapler to pad & Move pill bottle to pad \\
12 & Store Display & pick-place & Place object on stand & Rotate QR code \\
13 & Tabletop Arrange & pick-place & Place container on plate & Place fan on pad \\
14 & Workspace Setup & pick-place & Place mouse on pad & Place phone on stand \\
15 & Bakery Close & placement-to-interaction & Place bread in basket & Turn switch \\
16 & Kitchen Switch & placement-to-interaction & Place container on plate & Turn switch \\
17 & Room Comfort & placement-to-interaction & Place fan on pad & Turn switch \\
18 & Smart Home & placement-to-interaction & Place phone on stand & Turn switch \\
\bottomrule
\end{tabular}}
\end{table}

\begin{table}[p]
\centering
\caption{\textbf{Per-task atomic success rates on the RoboTwin benchmark.}
For each model, we report separate Clean and Rand. columns; Rand. denotes Randomized success rate (\%).
FastWAM results are from \emph{Fast-WAM: Do World Action Models Need Test-time Future Imagination?}~\citep{yuan2026fastwam};
LingBot-VLA results are from \emph{A Pragmatic VLA Foundation Model}~\citep{wu2026pragmatic};
Motus and $\pi_{0.5}$ results are from \emph{Motus: A Unified Latent Action World Model}~\citep{bi2025motus}.}
\label{tab:atomic_clean_randomized_success}
\scriptsize
\setlength{\tabcolsep}{2pt}
\renewcommand{\arraystretch}{1.0}
\begin{adjustbox}{width=0.80\textwidth,max totalheight=0.85\textheight}
\begin{tabular}{@{}lcccccccc@{}}
\toprule
Task & \multicolumn{2}{c}{FastWAM} & \multicolumn{2}{c}{Motus} & \multicolumn{2}{c}{LingBot-VLA} & \multicolumn{2}{c}{$\pi_{0.5}$} \\
\cmidrule(lr){2-3}\cmidrule(lr){4-5}\cmidrule(lr){6-7}\cmidrule(lr){8-9}
 & Clean & Rand. & Clean & Rand. & Clean & Rand. & Clean & Rand. \\
\midrule
Average & 91.88 & 91.78 & 88.66 & 87.02 & 86.50 & 85.34 & 82.74 & 76.76 \\
Adjust Bottle & 100 & 100 & 89 & 93 & 100 & 100 & 100 & 99 \\
Beat Block Hammer & 99 & 97 & 95 & 88 & 87 & 91 & 96 & 93 \\
Blocks Ranking RGB & 100 & 100 & 99 & 97 & 92 & 91 & 92 & 85 \\
Blocks Ranking Size & 94 & 98 & 75 & 63 & 66 & 73 & 49 & 26 \\
Click Alarmclock & 100 & 100 & 100 & 100 & 93 & 26 & 98 & 89 \\
Click Bell & 100 & 100 & 100 & 100 & 32 & 19 & 99 & 66 \\
Dump Bin Bigbin & 97 & 96 & 95 & 91 & 97 & 92 & 92 & 97 \\
Grab Roller & 100 & 100 & 100 & 100 & 100 & 99 & 100 & 100 \\
Handover Block & 95 & 81 & 86 & 73 & 80 & 83 & 66 & 57 \\
Handover Mic & 99 & 100 & 78 & 63 & 94 & 98 & 98 & 97 \\
Hanging Mug & 58 & 62 & 38 & 38 & 32 & 27 & 18 & 17 \\
Lift Pot & 100 & 100 & 96 & 99 & 100 & 99 & 96 & 85 \\
Move Can Pot & 90 & 88 & 34 & 74 & 79 & 84 & 51 & 55 \\
Move Pillbottle Pad & 100 & 99 & 93 & 96 & 93 & 94 & 84 & 61 \\
Move Playingcard Away & 100 & 100 & 100 & 96 & 96 & 99 & 96 & 84 \\
Move Stapler Pad & 77 & 64 & 83 & 85 & 74 & 49 & 56 & 42 \\
Open Laptop & 98 & 100 & 95 & 91 & 96 & 96 & 90 & 96 \\
Open Microwave & 62 & 45 & 95 & 91 & 91 & 75 & 34 & 77 \\
Pick Diverse Bottles & 80 & 85 & 90 & 91 & 79 & 86 & 81 & 71 \\
Pick Dual Bottles & 100 & 96 & 96 & 90 & 82 & 95 & 93 & 63 \\
Place A2B Left & 95 & 93 & 88 & 79 & 86 & 83 & 87 & 82 \\
Place A2B Right & 93 & 99 & 91 & 87 & 74 & 77 & 87 & 84 \\
Place Bread Basket & 91 & 93 & 91 & 94 & 92 & 93 & 77 & 64 \\
Place Bread Skillet & 90 & 93 & 86 & 83 & 90 & 89 & 85 & 66 \\
Place Burger Fries & 96 & 99 & 98 & 98 & 95 & 96 & 94 & 87 \\
Place Can Basket & 71 & 69 & 81 & 76 & 68 & 78 & 62 & 62 \\
Place Cans Plasticbox & 99 & 96 & 98 & 94 & 97 & 100 & 94 & 84 \\
Place Container Plate & 96 & 100 & 98 & 99 & 99 & 99 & 99 & 95 \\
Place Dual Shoes & 94 & 88 & 93 & 87 & 80 & 83 & 75 & 75 \\
Place Empty Cup & 100 & 100 & 99 & 98 & 100 & 100 & 100 & 99 \\
Place Fan & 96 & 96 & 91 & 87 & 91 & 79 & 87 & 85 \\
Place Mouse Pad & 83 & 89 & 66 & 68 & 82 & 78 & 60 & 39 \\
Place Object Basket & 89 & 88 & 81 & 87 & 90 & 91 & 80 & 76 \\
Place Object Scale & 90 & 97 & 88 & 85 & 84 & 90 & 86 & 80 \\
Place Object Stand & 90 & 94 & 98 & 97 & 97 & 93 & 91 & 85 \\
Place Phone Stand & 97 & 99 & 87 & 86 & 92 & 93 & 81 & 81 \\
Place Shoe & 96 & 99 & 99 & 97 & 99 & 94 & 92 & 93 \\
Press Stapler & 90 & 97 & 93 & 98 & 90 & 88 & 87 & 83 \\
Put Bottles Dustbin & 95 & 90 & 81 & 79 & 88 & 92 & 84 & 79 \\
Put Object Cabinet & 94 & 89 & 88 & 71 & 92 & 86 & 80 & 79 \\
Rotate QRcode & 93 & 89 & 89 & 73 & 93 & 84 & 89 & 87 \\
Scan Object & 89 & 92 & 67 & 66 & 91 & 97 & 72 & 65 \\
Shake Bottle & 100 & 100 & 100 & 97 & 99 & 100 & 99 & 97 \\
Shake Bottle Horizontally & 100 & 100 & 100 & 98 & 100 & 100 & 99 & 99 \\
Stack Blocks Three & 95 & 97 & 91 & 95 & 92 & 99 & 91 & 76 \\
Stack Blocks Two & 100 & 100 & 100 & 98 & 100 & 100 & 97 & 100 \\
Stack Bowls Three & 80 & 81 & 79 & 87 & 72 & 83 & 77 & 71 \\
Stack Bowls Two & 92 & 98 & 98 & 98 & 92 & 95 & 95 & 96 \\
Stamp Seal & 90 & 94 & 93 & 92 & 76 & 86 & 79 & 55 \\
Turn Switch & 61 & 59 & 84 & 78 & 61 & 65 & 62 & 54 \\
\bottomrule
\end{tabular}\end{adjustbox}
\end{table}

\section{HSE Details}
\label{sec:HSE_Details}

This section describes the implementation of \textsc{HSE}. \textsc{HSE} contains three executable components: task decomposition, progress predicate prediction, and subtask switching. The VLA policy itself is not modified by \textsc{HSE}; it is treated as an instruction conditioned atomic executor.

\subsection{VLA policy interface}

Let $\mathcal{Z}=\{\task{1},\ldots,\task{K}\}$ denote the atomic task vocabulary. At time $t$, the VLA policy receives an observation history
\begin{equation}
    h_t=o_{t-H+1:t}
\end{equation}
and one active atomic instruction $\ell\in\mathcal{Z}$. It predicts an action chunk of horizon $L$:
\begin{equation}
    a_{t:t+L-1}\sim \pi_\theta(\cdot\mid h_t,\ell).
\end{equation}
During \textsc{HSE} execution, the policy is never conditioned on multiple atomic instructions at the same invocation. The active instruction is selected by the subtask queue and may be updated only between two action chunk queries.

\subsection{Task decomposition}

The task decomposition module maps a composed language command into an ordered queue of atomic task instructions. We implement this module with prompt engineering and an LLM. The prompt provides the LLM with the closed atomic vocabulary $\mathcal{Z}$ and asks it to rewrite the composed command as an ordered list of valid atomic task names.

Given a composed command $c$, the decomposer outputs
\begin{equation}
    \hat q=\mathcal{D}(c)=(\hat q_1,\ldots,\hat q_m),
    \qquad \hat q_j\in\mathcal{Z}.
\end{equation}
For RoboCombine, each command contains two stages, so $m=2$. The output is constrained to use only task labels from the atomic vocabulary. If the raw LLM output contains extra text, we parse the ordered task names and discard non task tokens. This keeps the decomposition module as a lightweight language interface rather than a learned robot policy.

The role of this module is to decide \emph{what} atomic skill should be executed at each stage. In the predicted setting, \textsc{HSE} uses $\hat q$. In diagnostic settings, we replace it with the benchmark supplied queue $q^\star$ to remove decomposition errors and isolate later failure sources.

\subsection{Progress predictor for online handoffs}
\label{sec:progress_predictor_details}

In the progress based \textsc{HSE} mode, the handoff cue is produced by a lightweight progress predictor.
The predictor receives a single RGB frame from the \texttt{front\_camera} and the currently active atomic instruction.
We encode this frame instruction pair with a Florence based vision language encoder, then mean pool the encoder hidden states to obtain a 1024 dimensional feature vector.
This feature is passed to a temporal progress head consisting of a two layer GRU with hidden dimension 256, followed by an MLP with dimensions $256 \rightarrow 128 \rightarrow 1$ and a final sigmoid.

The output is a scalar progress score in $[0,1]$ for the current atomic stage.
It does not output robot actions, but only estimates whether the current subtask is close enough to completion to switch to the next atomic stage.
The recurrent head allows the score to integrate short temporal context rather than treating each frame independently.
When the score reaches the switching threshold, the predictor returns \texttt{triggered=True} and \textsc{HSE} advances to the next instruction in the subtask queue.
The subsequent call to \texttt{reset(instruction)} sets the new active atomic instruction, clears the GRU hidden state, and resets the internal step index. This software reset affects only the progress predictor and is distinct from the robot-reset oracle evaluated in Tab.~\ref{tab:prti_reset_controls}.
Consequently, the progress score rises within a stage, drops after a switch because the recurrent state and instruction are reset, and then rises again as the next stage progresses.

For training, we use cached Florence features and a tanh shaped progress target.
Let $p_t \in [0,1]$ denote the normalized frame index within an atomic segment.
The target is centered at \texttt{tanh\_center}$=0.5$ with slope \texttt{tanh\_slope}$=0.1$, producing a smooth low to high supervision signal over each stage.
The progress head is trained with a Huber loss with $\delta=0.1$, and the selected checkpoint is chosen by the validation metric \texttt{val\_acc@10\%}.
Thus, the deployed handoff module can be summarized as a Florence feature extractor plus a recurrent GRU progress head and a threshold trigger.

\subsection{Subtask switching}
The subtask switching module updates the active instruction according to the subtask queue and the progress predicate. Suppose the current subtask index is $j_k$ at the $k$th VLA invocation time $\tau_k$. The VLA policy is queried with the active instruction $q_{j_k}$:
\begin{equation}
    a_{\tau_k:\tau_k+L-1}
    \sim
    \pi_\theta(\cdot\mid h_{\tau_k},q_{j_k}).
\end{equation}

Because the VLA policy predicts an action chunk, \textsc{HSE} does not change the instruction inside the chunk. Instead, the progress predicate is monitored during the chunk and latched:
\begin{equation}
    G_k=\max_{\tau_k\leq s<\tau_k+L} g_s.
\end{equation}
The active subtask index is updated only before the next VLA query:
\begin{equation}
    j_{k+1}=\min\{j_k+G_k,m\}.
\end{equation}

This design makes switching compatible with chunk based VLA policies. Every predicted action chunk is conditioned on exactly one atomic instruction, while the next chunk may use the next instruction if the progress predicate fires. Importantly, the environment is not reset during switching. The downstream atomic policy must execute from the physical state produced by the previous subtask. This is why \textsc{HSE} directly tests whether atomic skills can be recombined across handoff states.

\section{PRTI Details}
\label{sec:PRTI_Details}

We implement \PRTI as an atomic post-training stage on top of a \(\pi_{0.5}\) Robotwin checkpoint. Unless otherwise specified, the model is initialized from an atomic skill checkpoint trained on the same task distribution, rather than from random initialization. We update all model parameters and keep the observation, instruction, and action interfaces unchanged from the backbone policy. The training data consist of a mixture of isolated atomic skill trajectories from two atomic data sources. The mixture contains on the order of 10K episodes, around 2M frames, and several hundred atomic task prompts. Importantly, all trajectories are single skill demonstrations. No complete composed trajectory is used for policy training. Each sample is conditioned only on the corresponding atomic task prompt.

The policy observes three RGB views, including one external camera and two wrist cameras, together with proprioceptive robot state. Images are resized to \(224 \times 224\), and the instruction input is provided as a natural language task prompt. For optimization, we perform distributed full parameter fine tuning over four accelerators. We train for 50K steps with a global batch size of 256, using AdamW, a cosine learning rate schedule with warmup, and an exponential moving average of model parameters. Other optimization settings follow standard VLA fine tuning practice.

This implementation keeps the training boundary atomic only. \PRTI changes the distribution of atomic initial states used during post training, but each trajectory is still supervised with a single atomic instruction. Therefore, the observed gains are not due to direct supervision on complete composed demonstrations, and are consistent with improved robustness of atomic skills when invoked from non canonical, handoff like states.

\subsection{Reset and initialization controls}
\paragraph{Random perturbation (Random).}
This control uses unstructured perturbations of initial states to test whether generic state diversity is sufficient to improve ordered execution. Such perturbations may break task-relevant physical correlations or produce states outside the realistic handoff distribution. In contrast, \PRTI samples trajectory-supported, physically feasible initial states from successful atomic demonstrations.

\paragraph{Robot-reset oracle (Reset).}
This execution-time control returns only the robot to its canonical pose between subtasks, leaving the object and contact states produced by the preceding subtask unchanged. It therefore tests whether restoring the robot pose alone is sufficient to bridge the start state gap. In real-world execution, this reset would require an additional collision-free return motion, which may itself disturb the scene. \PRTI inserts no such inter-subtask reset and executes the downstream skill directly from the complete handoff state.

Both controls are evaluated under oracle timed \textsc{HSE} over all 19 compositions with 20 rollouts each (380 episodes per method). As reported in Tab.~\ref{tab:prti_reset_controls}, Random achieves 16/380 ordered successes (4.21\%), Reset achieves 36/380 (9.47\%), and \PRTI achieves 69/380 (18.16\%).

\section{Real-World Details}
\label{sec:real_robot_deatils}
\subsection{Atomic tasks}

\paragraph{Real-robot atomic tasks.}
We instantiate 16 atomic manipulation tasks on a dual-arm AgileX robot platform with three camera views. These tasks cover drawer interaction, button pressing, object pick-and-place, and object transfer between the drawer and tabletop regions. The objects include an orange persimmon and a cube, and the workspace contains a drawer, left/right plates, center boards, and start/stop buttons. Each atomic task is defined by a natural-language prompt, and the full list of prompts is provided in Table~\ref{tab:real_world_task}. We evaluate each atomic task independently and report per-task success rates in Table~\ref{tab:atomic_success_rates}. These results characterize the reliability of the atomic skills that serve as the building blocks for our compositional real-robot experiments.

\newcolumntype{Y}{>{\raggedright\arraybackslash}X}
\newcolumntype{E}{>{\centering\arraybackslash}p{0.085\textwidth}}

\begin{table*}[t]
\centering
\caption{Atomic and compositional task prompts used in our experiments. The episode count is reported for each atomic task.}
\label{tab:task_prompts}
\footnotesize
\setlength{\tabcolsep}{3.5pt}
\renewcommand{\arraystretch}{1.16}

\begin{tabularx}{\textwidth}{@{}p{0.045\textwidth}Y E p{0.045\textwidth}Y E@{}}
\toprule
\rowcolor{gray!12}
\multicolumn{6}{@{}l}{\textbf{Atomic Tasks}} \\
\midrule
\textbf{ID} & \textbf{Task Prompt} & \textbf{Episodes}
& \textbf{ID} & \textbf{Task Prompt} & \textbf{Episodes} \\
\midrule
1  & Open the drawer. & 120
& 9  & Press the green start button at the bottom. & 40 \\

2  & Close the drawer. & 120
& 10 & Press the purple stop button at the top. & 40 \\

3  & Put the orange persimmon from the center plate onto the left plate. & 120
& 11 & Place the Cube from the left plate into the drawer. & 60 \\

4  & Put the orange persimmon from the center plate onto the right plate. & 120
& 12 & Place the Cube from the right plate into the drawer. & 60 \\

5  & Place the persimmon from the left plate into the drawer. & 120
& 13 & Move the Cube from the center board to the left board. & 60 \\

6  & Place the persimmon from the right plate into the drawer. & 120
& 14 & Move the Cube from the center board to the right board. & 60 \\

7  & Take the persimmon out of the drawer and place it on the left plate. & 120
& 15 & Take the Cube out of the drawer and place it on the left plate. & 60 \\

8  & Take the persimmon out of the drawer and place it on the right plate. & 120
& 16 & Take the Cube out of the drawer and place it on the right plate. & 60 \\
\bottomrule
\end{tabularx}

\vspace{0.75em}

\begin{tabularx}{\textwidth}{@{}p{0.15\textwidth}Y@{}}
\toprule
\rowcolor{gray!12}
\multicolumn{2}{@{}l}{\textbf{Compositional Tasks}} \\
\midrule
\textbf{ID} & \textbf{Task Prompt} \\
\midrule
1-7 &
Open the drawer, then take the persimmon out of the drawer and place it on the left plate. \\

1-8-6 &
Open the drawer, then take the persimmon out of the drawer and place it on the right plate, then place the persimmon from the right plate into the drawer. \\

1-15-11-16 &
Open the drawer, then take the Cube out of the drawer and place it on the left plate, then place the Cube from the left plate into the drawer, then take the Cube out of the drawer and place it on the right plate. \\

7-5-8-6-10 &
Take the persimmon out of the drawer and place it on the left plate, then place the persimmon from the left plate into the drawer, then take the persimmon out of the drawer and place it on the right plate, then place the persimmon from the right plate into the drawer, then press the purple stop button at the top. \\

1-12 &
Open the drawer, then place the Cube from the right plate into the drawer. \\

1-2 &
Open the drawer, then close the drawer. \\

1-6-2 &
Open the drawer, then place the persimmon from the right plate into the drawer, then close the drawer. \\

1-5-2-10 &
Open the drawer, then place the persimmon from the left plate into the drawer, then close the drawer, then press the purple stop button at the top. \\

4-1-6-2-10 &
Put the orange persimmon from the center plate onto the right plate, then open the drawer, then place the persimmon from the right plate into the drawer, then close the drawer, then press the purple stop button at the top. \\

3-10 &
Put the orange persimmon from the center plate onto the left plate, then press the purple stop button at the top. \\

4-10 &
Put the orange persimmon from the center plate onto the right plate, then press the purple stop button at the top. \\

9-3-10 &
Press the green start button at the bottom, then put the orange persimmon from the center plate onto the left plate, then press the purple stop button at the top. \\

9-14-10 &
Press the green start button at the bottom, then move the Cube from the center board to the right board, then press the purple stop button at the top. \\
\bottomrule
\end{tabularx}
\label{tab:real_world_task}
\end{table*}
\newcolumntype{C}{>{\centering\arraybackslash}p{0.045\textwidth}}

\begin{table*}[!t]
\centering
\caption{Atomic-task success rates of the $\pi_{0.5}$ model on the real robot.}
\label{tab:atomic_success_rates}
\scriptsize
\setlength{\tabcolsep}{2.2pt}
\renewcommand{\arraystretch}{1.15}

\begin{tabular}{@{}c C C C C C C C C C C C C C C C C@{}}
\toprule
\rowcolor{gray!12}
\textbf{Task ID}
& \textbf{1} & \textbf{2} & \textbf{3} & \textbf{4}
& \textbf{5} & \textbf{6} & \textbf{7} & \textbf{8}
& \textbf{9} & \textbf{10} & \textbf{11} & \textbf{12}
& \textbf{13} & \textbf{14} & \textbf{15} & \textbf{16} \\
\midrule
\textbf{Success}
& 10/10 & 9/10 & 8/10 & 7/10
& 5/10 & 8/10 & 6/10 & 7/10
& 10/10 & 8/10 & 6/10 & 5/10
& 9/10 & 7/10 & 3/10 & 2/10 \\
\bottomrule
\end{tabular}
\end{table*}

\subsection{Mean-progress metric}
\label{sec:real_robot_mean_progress}

For each method and compositional task, we average mean progress over five physical rollouts. Each rollout receives the score of the furthest ordered milestone it reaches under a fixed task-length-specific rubric. The same rubric is applied to all methods, and the milestones capture both atomic-subtask completion and successful handoff to the next subtask. Table~\ref{tab:real_robot_progress_rubric} gives the complete scoring protocol.

\begin{table*}[t]
    \centering
    \caption{Ordered-milestone rubric for real-robot mean progress. For the five-stage tasks, each pair gives the score for initiating/completing the corresponding subtask.}
    \label{tab:real_robot_progress_rubric}
    \small
    \setlength{\tabcolsep}{5pt}
    \renewcommand{\arraystretch}{1.12}
    \begin{tabular}{@{}p{0.12\textwidth}p{0.82\textwidth}@{}}
        \toprule
        \textbf{Task length} & \textbf{Ordered milestones and scores} \\
        \midrule
        2 stages & Initiate $S_1$: 0.2; complete $S_1$: 0.5; switch to $S_2$: 0.7; complete $S_2$: 1.0. \\
        3 stages & Initiate $S_1$: 0.1; complete $S_1$: 0.3; switch to $S_2$: 0.4; complete $S_2$: 0.6; switch to $S_3$: 0.7; complete $S_3$: 1.0. \\
        4 stages & Initiate $S_1$: 0.1; complete $S_1$: 0.2; switch to $S_2$: 0.3; complete $S_2$: 0.4; switch to $S_3$: 0.5; complete $S_3$: 0.7; switch to $S_4$: 0.8; complete $S_4$: 1.0. \\
        5 stages & Initiate/complete $S_1,\ldots,S_5$: 0.1/0.2, 0.3/0.4, 0.5/0.6, 0.7/0.8, and 0.9/1.0, respectively. \\
        \bottomrule
    \end{tabular}
\end{table*}

\subsection{Visualization}

\begin{figure*}[t]
    \centering
    \includegraphics[width=0.98\linewidth]{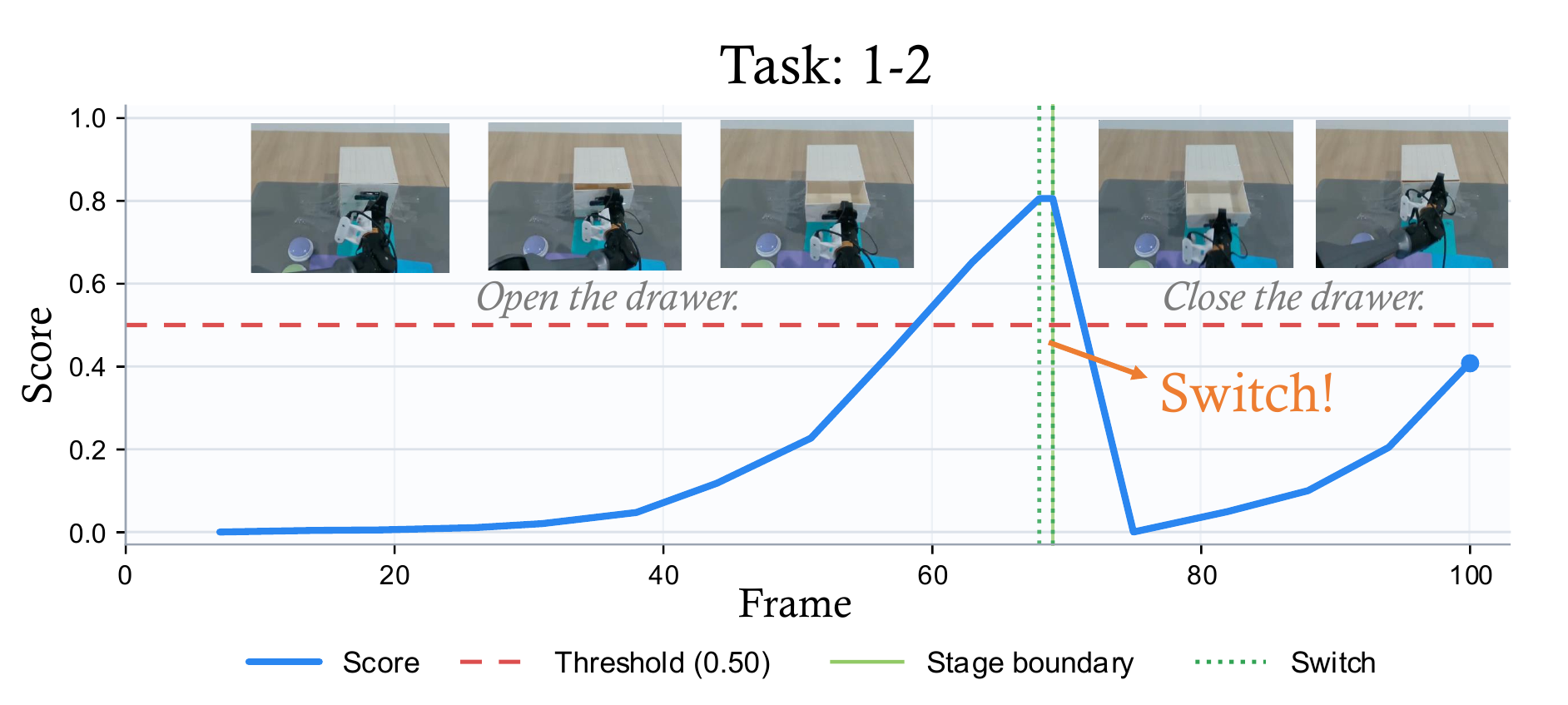}
    \caption{\textbf{Real-robot execution trace on the two-step compositional task \textit{1-2}}. The top row shows representative frames from the rollout, and the blue curve plots the progress predictor score over time. The red dashed line indicates the switching threshold, green vertical lines mark stage boundaries, and green dotted lines indicate the moments when the system switches to the next subtask. The score rises within each stage and drops after a successful switch, illustrating hierarchical execution over the full task sequence.}
    \label{fig:real_2task}
\end{figure*}

\begin{figure*}[t]
    \centering
    \includegraphics[width=0.98\linewidth]{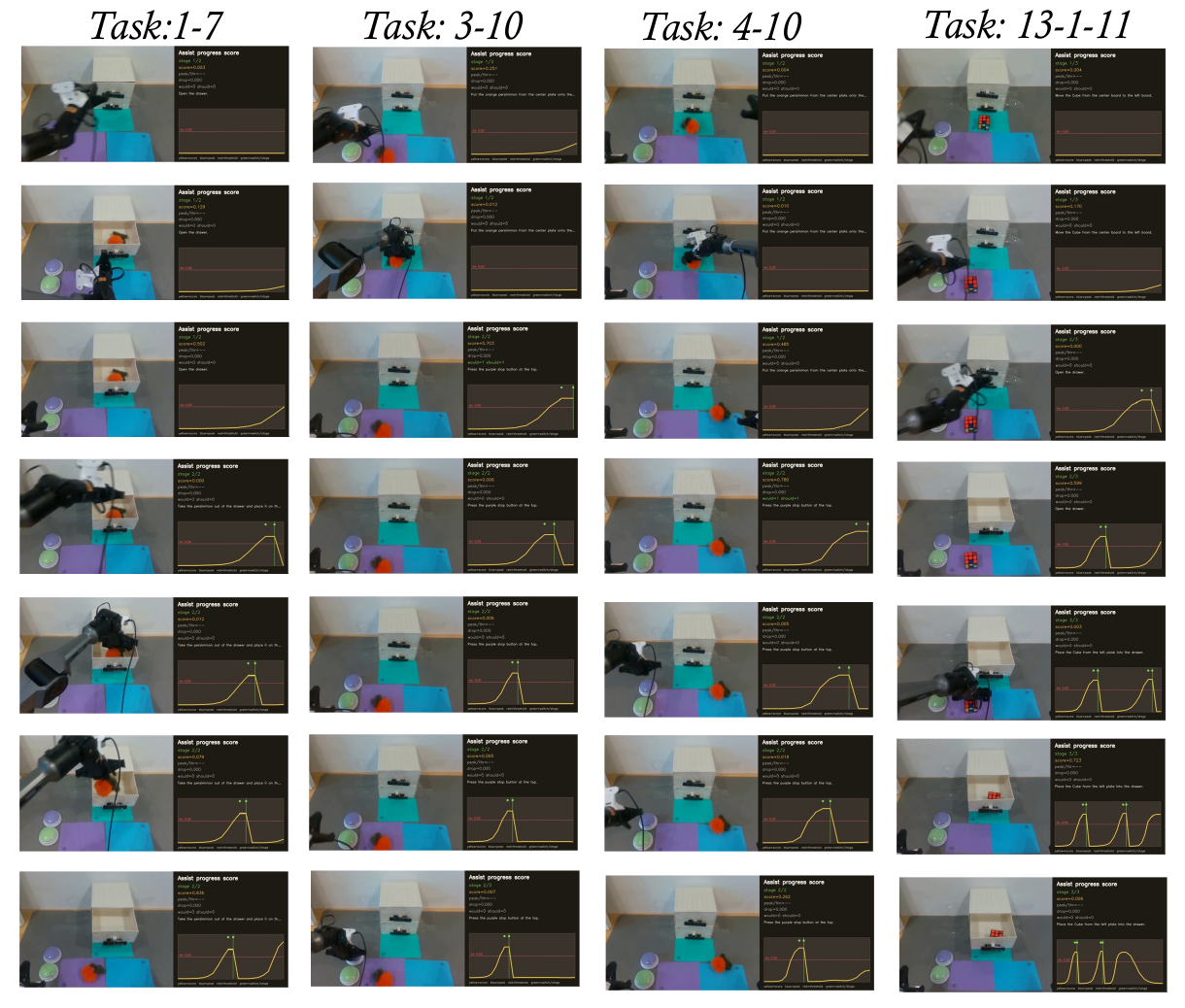}
    \caption{\textbf{Representative real-robot executions of our framework on four compositional tasks.} Each column corresponds to one task, and each row shows temporally sampled frames from the rollout. For each frame, the left panel shows the real-robot observation, while the right panel visualizes the progress predictor score over frames. The score curves provide an interpretable trace of subtask progress and switching behavior during execution, illustrating how our framework completes multi-step tasks by sequentially invoking atomic skills.}
    \label{fig:real_vis}
\end{figure*}

We further visualize the real-robot execution process to examine whether the progress predictor provides meaningful subtask-level signals during deployment.
Fig.~\ref{fig:real_2task} shows a detailed execution trace for the two-step task \texttt{1-2}. The progress score increases as the robot completes the current subtask, crosses the switching threshold near the end of the first stage, and then drops after switching to the next subtask. This pattern indicates that the predictor captures task progress rather than merely producing static confidence scores. Fig.~\ref{fig:real_vis} provides additional examples across four compositional tasks with different task lengths and object interactions. Across these rollouts, the score curves generally form stage-wise peaks aligned with subtask completion and switching events. These visualizations illustrate that our framework can monitor subtask progress online and use the resulting signals to coordinate sequential atomic-skill execution on the real robot.

\end{document}